\documentclass[pdflatex,sn-mathphys-num]{sn-jnl}

\usepackage{graphicx}%
\usepackage{multirow}%
\usepackage{amsmath,amssymb,amsfonts}%
\usepackage{amsthm}%
\usepackage{mathrsfs}%
\usepackage[title]{appendix}%
\usepackage{xcolor}%
\usepackage{textcomp}%
\usepackage{manyfoot}%
\usepackage{booktabs}%
\usepackage{algorithm}%
\usepackage{algorithmicx}%
\usepackage{algpseudocode}%
\usepackage{listings}%
\usepackage{tabularx}
\usepackage{makecell}

\theoremstyle{thmstyleone}%
\theoremstyle{thmstyletwo}%

\theoremstyle{thmstylethree}%

\begin{document}

\title[Article Title]{A Survey for Large Language Models in Biomedicine
}


\author[1,2,3]{\fnm{Chong} \sur{Wang}} \equalcont{These authors contributed equally to this work.}

\author[1]{\fnm{Mengyao} \sur{Li}} \equalcont{These authors contributed equally to this work.}

\author[4]{\fnm{Junjun} \sur{He}} \equalcont{These authors contributed equally to this work.}

\author[5]{\fnm{Zhongruo} \sur{Wang}}

\author[6,7]{\fnm{Erfan} \sur{Darzi}}

\author[8]{\fnm{Zan} \sur{Chen}}

\author[4,9]{\fnm{Jin} \sur{Ye}}

\author[4]{\fnm{Tianbin} \sur{Li}}

\author[4]{\fnm{Yanzhou} \sur{Su}}

\author[10,11]{\fnm{Jing} \sur{Ke}}


\author[1]{\fnm{Kaili} \sur{Qu}}

\author[1]{\fnm{Shuxin} \sur{Li}}

\author[1]{\fnm{Yi} \sur{Yu}}

%

\author[13]{\fnm{Pietro} \sur{Liò}}

\author*[14]{\fnm{Tianyun} \sur{Wang}}

\author*[4,8,15,16]{\fnm{Yu Guang} \sur{Wang}}

\author*[17]{\fnm{Yiqing} \sur{Shen}}
\email{yshen92@jhu.edu}

\affil[1]{\orgdiv{School of Medical Engineering}, \orgname{Xinxiang Medical University}, \orgaddress{ \city{Xinxiang},  \country{China}}}

\affil[2]{\orgdiv{Engineering Technology Research Center of Neurosense and Control of Henan Province},\orgaddress{ \city{Xinxiang},  \country{China}}}

\affil[3]{\orgdiv{Henan International Joint Laboratory of Neural Information Analysis and Drug Intelligent Design}, \orgaddress{ \city{Xinxiang}, \country{China}}}

\affil[4]{\orgname{Shanghai AI Laboratory}, \orgaddress{\city{Shanghai}, \country{China}}}

\affil[5]{\orgname{Amazon}, \orgaddress{\city{Palo Alto, CA}, \country{USA}}}

\affil[6]{\orgname{Boston Children's Hospital}, \orgaddress{\city{MA}, \country{USA}}}

\affil[7]{\orgdiv{Harvard Medical School}, \orgname{Harvard University}, \orgaddress{\city{MA}, \country{USA}}}

\affil[8]{\orgname{Toursun Synbio}, \orgaddress{\city{Shanghai}, \country{China}}}

\affil[9]{\orgdiv{Department of Data Science \& AI, Faculty of IT}, \orgname{Monash University}, \orgaddress{\city{Melbourne},  \country{Australia}}}

\affil[10]{\orgdiv{School of Electronic Information and Electrical Engineering}, \orgname{Shanghai Jiao Tong University}, \orgaddress{\city{Shanghai}, \country{China}}}

\affil[11]{\orgdiv{School of Computer Science and Engineering}, \orgname{University of New South Wales}, \orgaddress{\city{Sydney}, \country{Australia}}}

\affil[12]{\orgdiv{School of Life Sciences and Biotechnology}, \orgname{Shanghai Jiao Tong University}, \orgaddress{\city{Shanghai}, \country{China}}}

\affil[13]{\orgdiv{Department of Computer Science and Technology}, \orgname{University of Cambridge}, \orgaddress{\city{Cambridge}, \country{UK}}}

\affil[14]{\orgdiv{School of Basic Medical Sciences}, \orgname{Xinxiang Medical University}, \orgaddress{ \city{Xinxiang}, \country{China}}}

\affil[15]{\orgdiv{Institute of Natural Sciences}, \orgname{Shanghai Jiao Tong University}, \orgaddress{\city{Shanghai}, \country{China}}}

\affil[16]{\orgdiv{School of Mathematics and Statistics}, \orgname{University of New South Wales}, \orgaddress{\city{Sydney}, \country{Australia}}}

\affil[17]{\orgdiv{Department of Computer Science}, \orgname{Johns Hopkins University}, \orgaddress{\city{MD}, \country{USA}}}


\abstract{
Recent breakthroughs in large language models (LLMs) offer unprecedented natural language understanding and generation capabilities. 
However, existing surveys on LLMs in biomedicine often focus on specific applications or model architectures, lacking a comprehensive analysis that integrates the latest advancements across various biomedical domains.
This review, based on an analysis of 484 publications sourced from databases including PubMed, Web of Science, and arXiv, provides an in-depth examination of the current landscape, applications, challenges, and prospects of LLMs in biomedicine, distinguishing itself by focusing on the practical implications of these models in real-world biomedical contexts.
Firstly, we explore the capabilities of LLMs in zero-shot learning across a broad spectrum of biomedical tasks, including diagnostic assistance, drug discovery, and personalized medicine, among others, with insights drawn from 137 key studies.
Then, we discuss adaptation strategies of LLMs, including fine-tuning methods for both uni-modal and multi-modal LLMs to enhance their performance in specialized biomedical contexts where zero-shot fails to achieve, such as medical question answering and efficient processing of biomedical literature. 
Finally, we discuss the challenges that LLMs face in the biomedicine domain including data privacy concerns, limited model interpretability, issues with dataset quality, and ethics due to the sensitive nature of biomedical data, the need for highly reliable model outputs, and the ethical implications of deploying AI in healthcare.
To address these challenges, we also identify future research directions of LLM in biomedicine including federated learning methods to preserve data privacy and integrating explainable AI methodologies to enhance the transparency of LLMs. 
As this field of LLM rapidly evolves, continued research and development are essential to fully harness the capabilities of LLMs in biomedicine while ensuring their responsible and effective deployment.
}

\maketitle
\section{Introduction}

\begin{figure}[!ht]
    \centering
    \includegraphics[width=\linewidth]{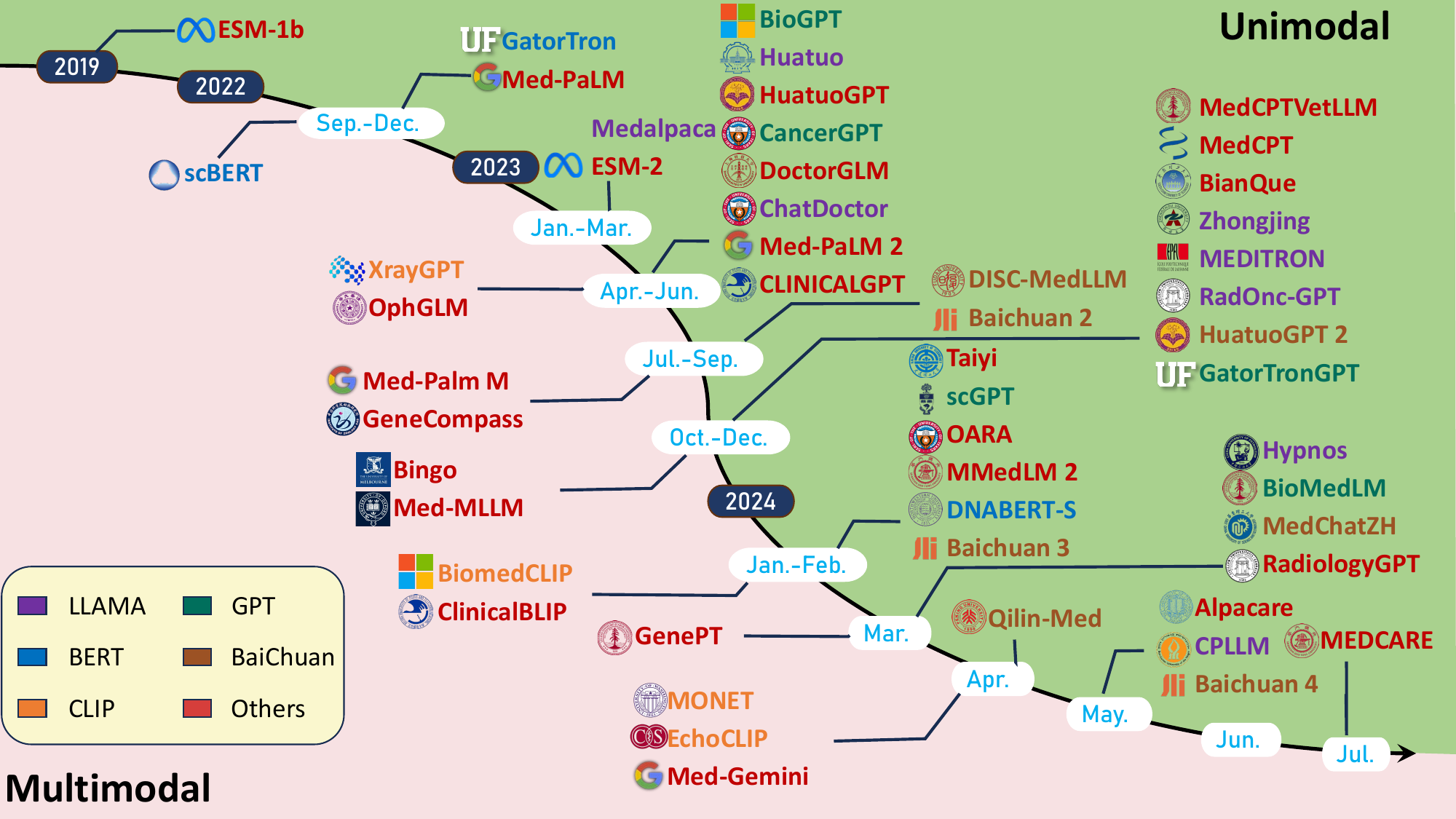}
    \caption{Chronological overview of LLMs and their variants in biomedical applications from 2019 to 2024. The timeline illustrates the evolution of both unimodal (top) and multimodal (bottom) models, highlighting key developments across different model architectures including LLAMA, GPT, BERT, BaiChuan, CLIP, and others. Notable milestones such as ESM-1b, Med-PaLM, and BioGPT are shown, demonstrating the progress and diversification of LLMs in the biomedical domain.}
    \label{fig:overview}
\end{figure}

%
General-purpose large language models (LLMs) such as PaLM~\cite{chowdhery2023palm}, LLaMA~\cite{touvron2023llama,touvron2023llama2}, and the GPT series~\cite{brown2020language,achiam2023gpt} have demonstrated their versatility across a wide range of tasks.
These models excel in complex language understanding and generation tasks, including translation, summarization, and nuanced question answering~\cite{naveed2023comprehensive}.
The advancements in LLM capabilities can be largely attributed to the evolution of deep learning algorithms, particularly the introduction and subsequent optimization of the Transformer architecture~\cite{hadi2023large}. 
As LLMs continue to mature, their potential applications across various domains are becoming increasingly apparent, with the biomedical field emerging as a particularly promising area of impact.
Fig.~\ref{fig:overview} presents a chronological overview of LLM development and its variants in biomedical applications from 2019 to 2024.
This timeline illustrates the rapid evolution of both unimodal and multimodal LLMs.
Notable achievements in biomedical LLMs showcase the breadth and depth of their impact.
For instance, MedPaLM~\cite{singhal2023towards} has attained a 92.9\% agreement with clinical experts in providing detailed medical answers and reaching scientific consensus.
In the realm of genomics, scBERT~\cite{yang2022scbert} generates embeddings for each gene using an improved Performer architecture, enhancing the analysis of single-cell genomic data. 
The development of domain-specific LLMs like HuatuoGPT~\cite{zhang2023huatuogpt}, ChatDoctor~\cite{li2023chatdoctor}, and BenTsao~\cite{wang2023huatuo} demonstrates the capability for reliable medical dialogue, showcasing the potential of LLMs in clinical communication and decision support.
%
%
The progression from predominantly unimodal LLMs to an increasing number of multimodal LLM approaches reflects the growing adaptability of LLMs in addressing complex biomedical challenges. 
This shift enables the integration of diverse data types, such as text, images, and structured clinical data.

\begin{figure}[!ht]
    \centering
    \includegraphics[width=\linewidth]{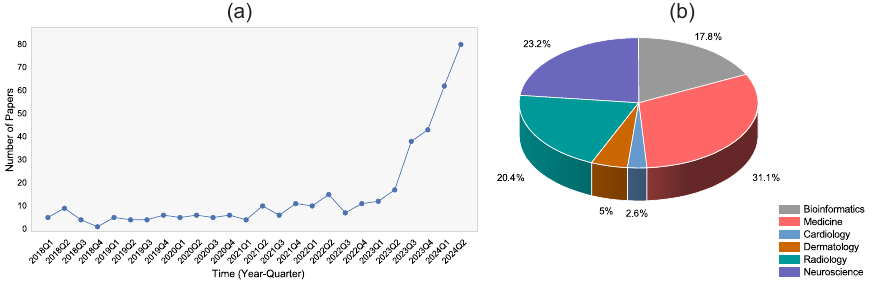}
    \caption{Trends and distribution of LLM research papers in biomedical fields from 2018 to 2024. (a) Temporal analysis of LLM research papers, showing quarterly publication counts. A surge in publications is evident beginning in 2021, reflecting growing interest and investment in applying LLMs to biomedical challenges. (b) Distribution of LLM research papers across biomedical specialties. Medicine (31.1\%) and Neuroscience (23.2\%) emerge as the dominant areas, followed by Radiology (20.4\%) and Bioinformatics (17.8\%). This distribution illustrates the broad applicability of LLMs across various medical domains and highlights potential areas for future development.}
    \label{fig:trend}
\end{figure}

The rapid growth and diversification of LLM research in biomedicine are further evidenced by the trends shown in Fig.~\ref{fig:trend}. 
A temporal analysis of LLM research papers in biomedical fields from 2018 to 2024 reveals an increase in publications, with a surge beginning in 2021 (Fig.~\ref{fig:trend}a).
This trend underscores the growing interest and investment in applying LLMs to biomedical challenges, reflecting both the technological advancements and the recognition of LLMs' potential to address healthcare and research needs.
The distribution of these research papers across various biomedical fields highlights `medicine' and `neuroscience' as the dominant areas of focus (Fig.~\ref{fig:trend}b).
This distribution demonstrates the broad applicability of LLMs across different medical specialties and research domains, while also indicating potential areas for future expansion and development.

The biomedical field encompasses a vast array of disciplines, from fundamental biological research to complex clinical applications, each characterized by specialized terminology and a evolving knowledge base~\cite{zhou2023survey}. 
This breadth and depth present challenges for the application of LLMs in biomedicine. 
The continuous influx of new research findings, treatment modalities, and pharmaceutical developments demands models capable of adapting to and integrating novel information swiftly~\cite{topol2019high}. 
Moreover, the high-stakes nature of biomedical applications necessitates an exceptionally high standard of accuracy and reliability from LLMs, which is a benchmark that current models may not consistently meet~\cite{sallam2023chatgpt,xie2023faithful}.
This shortcoming stems from the general-purpose nature of many LLMs, which can lead to misinterpretations and inference biases when confronted with the nuanced, context-dependent language of biomedical texts~\cite{peng2023study}.  
Furthermore, the field's reliance on sensitive patient data introduces additional complexities, requiring strict adherence to data protection and privacy regulations, which poses both technical and ethical challenges in implementation~\cite{mumtaz2024llms}.
Despite these hurdles, the potential for LLM applications in biomedicine remains promising. 
Models like BioMedLM~\cite{bolton2024biomedlm} demonstrate the capacity to accelerate scientific insight acquisition, while methods such as BianQue~\cite{chen2023bianque} and DISC-MedLLM~\cite{bao2023disc} show potential in providing medical advice during patient consultations, potentially alleviating clinical workloads.
However, the widespread adoption of these applications hinges on specialized training and optimization of LLMs to enhance their reliability and specificity in biomedical contexts.

While several surveys have explored the applications of LLMs in biomedicine, our review stands out due to its comprehensive scope and interdisciplinary approach. Unlike previous surveys that often focused on specific applications or model architectures, we provide an in-depth analysis of LLMs across various biomedical fields, ranging from genomics to clinical practice. Covering the period from 2019 to 2024, we offer insights into the latest developments and future trends, including both unimodal and multimodal LLM approaches. This review is based on an analysis of 484 publications from multiple databases, providing a thorough examination of the current state, applications, challenges, and prospects of LLMs in biomedicine. We evaluate the zero-shot performance of LLMs across various biomedical tasks, analyze adaptation strategies for both unimodal and multimodal approaches, and identify specific challenges faced by LLMs in biomedical applications, proposing potential solutions. By exploring the potential impact of LLMs on medical practice, biomedical research, and healthcare systems, our goal is to provide researchers, healthcare professionals, and policymakers with a clear roadmap to understand and leverage LLMs in biomedicine, facilitating informed decision-making and guiding future research efforts.

%

\section{Background}

Through extensive pre-training and fine-tuning, LLMs are capable of learning and capturing complex patterns and semantic relationships within language. 
%
%
In the following sections, we provide a detailed overview of the core structures of LLMs, their common model architectures, and fine-tuning techniques. 
The design of LLMs typically relies on the Transformer architecture and can be categorized into three main types: encoder-only, decoder-only, and encoder-decoder~\cite{minaee2024large}. 
Each architecture has distinct advantages and is suited for different types of tasks.

\subsection{Encoder-Only Architecture}
Encoder-only models focus on understanding and representing input text~\cite{devlin2018bert}.
These models are particularly adept at tasks that require deep contextual understanding, such as text classification, named entity recognition, and sentiment analysis.
The Bidirectional Encoder Representations from Transformers (BERT)~\cite{devlin2018bert} is an example of this architecture.
BERT's key innovation is its bidirectional nature, allowing it to capture context from both left and right sides of each word in a sentence.
This bidirectional encoding provides a richer representation of text compared to previous unidirectional models.
BERT achieves this through its ``masked language model'' pre-training objective, where the model learns to predict randomly masked words in a sentence, forcing it to consider the full context.
Another notable encoder-only model is the Contrastive Language-Image Pretraining (CLIP) model~\cite{radford2021learning}. 
CLIP extends the encoder architecture to multimodal learning, integrating both text and image inputs.
By using contrastive learning, CLIP learns to align textual and visual representations in a shared embedding space.
The application of encoder-only models has achieved significant advancements in specialized scientific domains, particularly in the biomedical field. Notable examples include scBERT~\cite{yang2022scbert}, which generates fine-grained gene embeddings to process biomedical data, demonstrating exceptional performance in genomic analysis. Another prominent model, BioBERT~\cite{lee2020biobert}, is specifically designed for biomedical text mining, enhancing tasks such as named entity recognition and relation extraction within scientific literature. These specialized adaptations highlight the versatility of encoder-only models in addressing complex biomedical challenges.

\subsection{Decoder-Only Architecture}

Decoder-only models are designed for generative tasks, producing output sequences from left to right.
These models excel in text generation, dialogue systems, and creative writing applications. 
The Generative Pre-trained Transformer (GPT) series, culminating in the recent GPT-4, exemplifies this architecture~\cite{brown2020language,achiam2023gpt} with a unidirectional decoder structure, predicting each token based on the preceding context.
This approach allows for coherent and contextually appropriate text generation. 
The GPT models are trained on vast corpora of text, enabling them to capture complex language patterns and generate human-like text across diverse domains.
Other notable decoder-only models include LLaMA~\cite{touvron2023llama} and PaLM~\cite{chowdhery2023palm}. 
These models have optimized the decoder architecture for improved efficiency and scalability.
LLaMA, for instance, demonstrates strong performance with fewer parameters than its predecessors, while PaLM showcases improved multitask learning capabilities across various NLP benchmarks.
Decoder-only architectures have also been extended to multimodal applications. DALL·E~\cite{ramesh2021zero}, for example, uses a decoder to generate images from textual descriptions.
In the biomedical domain, decoder-only models have shown promising applications. For instance, they have been adapted for medical report generation and drug discovery tasks, such as BioGPT~\cite{luo2022biogpt}, CancerGPT~\cite{li2024cancergpt} and Med-PaLM~\cite{singhal2023large}.

\subsection{Encoder-Decoder Architecture}
The encoder-decoder architecture, also known as the sequence-to-sequence (seq2seq) model, combines the strengths of both encoder and decoder components. 
This design makes it suitable for tasks that involve transforming one sequence into another, such as machine translation, text summarization, and question answering.
In this architecture, the encoder processes the input sequence and compresses it into a latent representation. 
The decoder then uses this representation to generate the target sequence~\cite{du2021glm}.
This separation of encoding and decoding allows the model to handle input and output sequences of different lengths and structures effectively.
Two examples of encoder-decoder models are the Text-To-Text Transfer Transformer (T5)~\cite{raffel2020exploring} and Bidirectional and Auto-Regressive Transformers (BART)~\cite{lewis2019bart}
T5 adopts a unified approach by framing all NLP tasks as text-to-text problems, demonstrating remarkable versatility and strong multitask processing capabilities.
BART, on the other hand, combines the bidirectional nature of BERT-like encoders with the autoregressive generation of GPT-like decoders, making it particularly effective for text generation and repair tasks.
%
%
In biomedical applications, encoder-decoder models have shown significant potential. For instance, BioBART~\cite{yuan2022biobart} has been adapted for biomedical text generation and summarization tasks. Another notable example is GeneCompass~\cite{yang2023genecompass}, a cross-species large language model designed to decipher gene regulatory mechanisms. These applications highlight the architecture's versatility in addressing complex biomedical challenges, from text processing to unraveling the intricacies of genetic regulation across different species.

\section{LLMs in Zero-Shot Biomedical Applications}

\begin{figure}[!ht]
    \centering
    \includegraphics[width=\linewidth]{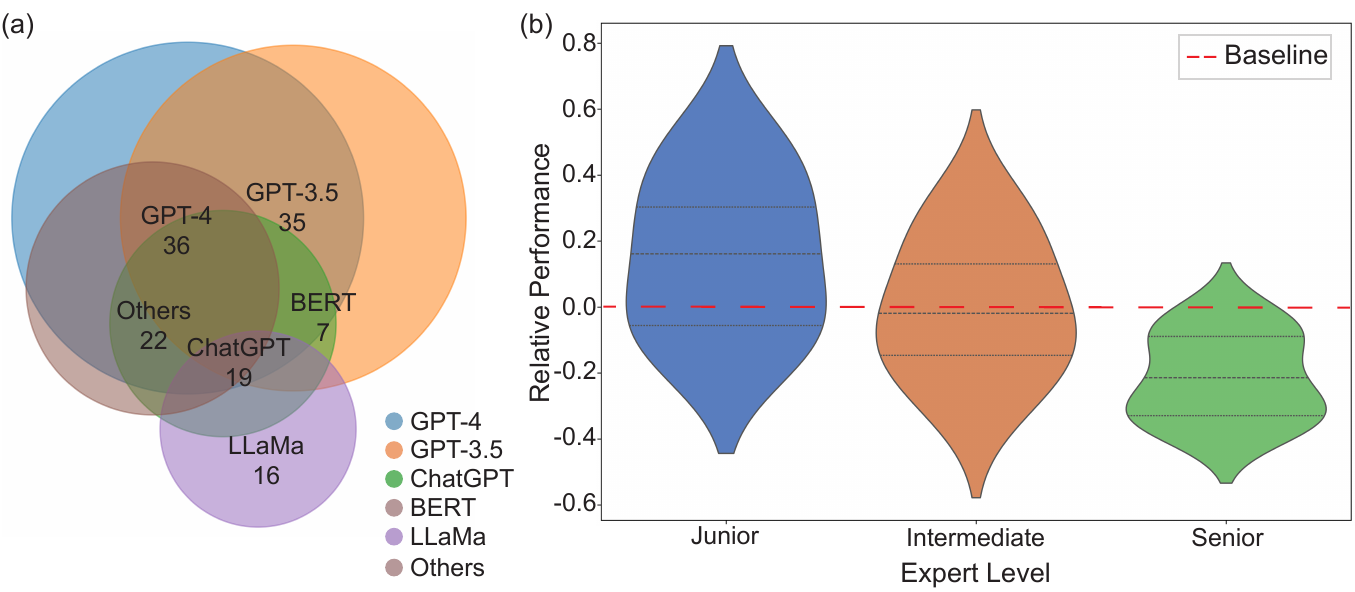}
    \caption{Evaluation of LLMs in biomedical applications in a zero-shot manner.
(a) Venn diagram illustrating the distribution and overlap of studies evaluating various LLMs (GPT-4, GPT-3.5, ChatGPT, BERT, LLaMA, and others) in zero-shot biomedical tasks. The numbers indicate the frequency of studies for each model.
(b) Violin plots comparing the relative performance of LLMs across different levels of biomedical expertise (Junior, Intermediate, Senior) against a baseline. The y-axis represents relative performance, with positive values indicating superior performance and negative values indicating inferior performance compared to the baseline. The width of each plot reflects the distribution of performance at each expertise level.}
    \label{fig:zero-shot}
\end{figure}

The potential of general-purpose LLMs has generated considerable interest in the biomedical field. 
Fig. \ref{fig:zero-shot}a illustrates the distribution of studies evaluating various LLMs in zero-shot biomedical tasks. 
GPT-4 and GPT-3.5 are the most frequently studied models, with 36 and 35 studies respectively, followed by ChatGPT with 19 studies.
This distribution highlights the current focus on OpenAI's models in biomedical research, with overlap between studies of different models indicating a trend towards comparative analysis.
Despite the performance of these LLMs across various domains, their efficacy in addressing the unique challenges of the biomedical field remains uncertain.
The specialized nature of biomedical terminology and the necessity to integrate specific clinical contexts pose challenges for these LLMs. 
To address this question, numerous studies have investigated the direct application of general-purpose LLMs in various biomedical disciplines, focusing on their performance in clinical diagnosis, decision support, drug development, genomics, personalized medicine, and biomedical literature analysis as elaborated in this section~\cite{sallam2023chatgpt,li2024chatgpt,liu2023utility}. 
%
%

\subsection{Diagnostic Assistance}
Diagnostic assistance is a biomedical technology that encompasses clinical diagnosis and decision support~\cite{wu2022application}. 
It analyzes patients' clinical data and symptoms, integrates medical knowledge with algorithmic processing, and provides recommendations to aid physicians in disease diagnosis and treatment decisions\cite{kumar2023artificial}. 
It aims to enhance diagnostic accuracy and efficiency, helping doctors better understand patients' conditions and formulate personalized treatment plans. 
To evaluate the zero-shot capabilities of general-purpose LLMs in biomedical diagnosis, researchers have designed a series of questions across various specialties.
Studies have assessed LLM performance in oncology \cite{haver2023appropriateness,zhu2023can}, emergency medicine \cite{bushuven2023chatgpt}, ophthalmology \cite{mihalache2023performance,hu2023can}, and nursing \cite{kothari2023chatgpt}, with results indicating that LLMs can achieve accuracy levels comparable to those of human experts in diagnostic tasks across these domains. 
Ward \textit{et al.} \cite{ward2022quantitative} conducted a comparative study of LLM performance in neurosurgical scenarios.
They created 30 clinical scenarios with consensus-based key points for answers and invited physicians of varying experience levels to respond to diagnostic questions. The results showed that GPT-4 achieved 100\% accuracy in triage and diagnosis, while GPT-3.5 had an accuracy rate of 92.59\%. These results highlight GPT-4's exceptional diagnostic accuracy, underscoring its potential as a reliable tool in clinical decision-making.
In oncology, Deng \textit{et al.}\cite{deng2024evaluation} found that GPT-4 achieved a 100\% accuracy rate in triage and diagnosis across breast cancer clinical scenarios, aligning closely with senior medical professionals. Similarly, Haver \textit{et al.} \cite{haver2023appropriateness} demonstrated GPT-4's effectiveness in neurosurgery, where it achieved 100\% accuracy in diagnosing and triaging neurosurgical cases, with perfect sensitivity and specificity. These findings highlight GPT-4's growing potential as a reliable tool in clinical decision-making across various medical fields.

\subsection{Biomedical Omics and Drug Discovery}
Biomedical science is an interdisciplinary field that encompasses drug development, genomics, and protein research, among other areas~\cite{he2023chat,blanco2023role}.
It integrates engineering, biology, and medicine, utilizing advanced biotechnology techniques to study disease prevention, diagnosis, and treatment~\cite{houssein2019bmc}.
By exploring the molecular mechanisms of life processes, this field aims to develop novel biomedical approaches and pharmaceuticals to enhance human health and disease management.
For instance, one study harnessed a LLM for candidate gene prioritization and selection, significantly improving the efficiency of identifying potential gene-disease associations. This approach utilized advanced natural language processing techniques to analyze vast amounts of genetic and biomedical data, leading to the prioritization of genes with a strong likelihood of being implicated in specific diseases~\cite{shen2024harnessing}. In another study, BERT was utilized to identify drug-target interactions from the entire PubMed database, achieving an accuracy of 99\% and identifying 0.6 million new articles with relevant data~\cite{aldahdooh2022bert}. Furthermore, Hou \textit{et al.}~\cite{hou2024assessing} leveraged GPT-4 for cell type annotation in single-cell RNA-seq analysis, demonstrating that GPT-4 can accurately annotate cell types using marker gene information. This approach achieved over 75\% agreement with manual annotations in most studies and tissues, highlighting its potential to reduce the labor and expertise required for cell type annotation. Collectively, these advancements underscore the potential of AI-driven models to transform biomedical research, offering more precise and efficient tools for disease understanding and treatment development.

\subsection{Personalized Medicine}
LLMs have also demonstrated potential in democratizing medical knowledge through online medical consultations \cite{zhu2023can,haupt2023ai,howard2023chatgpt,zhang2024exploring}. 
This capability ensures broad accessibility to biomedical information and enables personalized customization based on individual conditions, which could have profound implications for telemedicine \cite{sallam2023chatgpt,shea2023use}.
However, the development of personalized treatment plans using LLMs requires strict adherence to medical ethics and patient privacy. 
It is important to ensure that all data collection, storage, and usage comply with legal regulations and ethical standards. 
Ferrario \textit{et al.} \cite{ferrario2024large} evaluated GPT-4's performance in responding to various medical ethics cases. 
Their findings indicated that while GPT-4 can identify and articulate complex medical ethical issues, it requires improvement in encoding real-world ethical dilemmas more deeply.
Sandmann \textit{et al}. \cite{sandmann2024systematic} conducted an assessment of LLMs in clinical decision-making. 
They evaluated the clinical accuracy of initial diagnoses, examination steps, and treatments for 110 cases across different clinical disciplines using ChatGPT, LLaMA, and a naive baseline. 
Their results showed that GPT-4 performed the best among the tested models. 
Importantly, this study suggests that open-source LLMs may offer a viable solution for addressing data privacy concerns in personalized medicine applications.

\subsection{Biomedical Literature and Research}
The integration of LLMs with biomedical research and writing has enhanced research efficiency, impartiality, and accessibility~\cite{meng2024application}.
This synergy allows experts and researchers to more effectively obtain, understand, and apply the latest biomedical information, thereby increasing research productivity. 
LLMs have demonstrated utility in multiple key areas of biomedical literature, including literature retrieval, outline preparation, abstract writing, and translation tasks.
Mojadeddi \textit{et al.} \cite{mojadeddi2023impact} evaluated ChatGPT's performance in article writing. 
Their findings indicated that while ChatGPT can expedite the writing process, it has not yet reached the level of professional biomedical writers and has certain limitations. 
This underscores the need for further investigation into AI capabilities in scientific writing.
Huespe~\cite{huespe2023clinical} assessed GPT-3.5's ability to write the background section of critical care clinical research questions. 
In this study, 80 researchers were invited to distinguish between human-written and LLM-generated content. The results suggested that GPT-3.5's writing ability is comparable to that of biomedical researchers in this specific task.

\subsection{Benchmark Datasets and Evaluation Metrics}
A variety of benchmark datasets have been utilized in the evaluation on the performance of LLMs to biomedical inquiries. 
Table \ref{tab:benchmark_datasets} presents benchmark datasets used in recent studies. 
These datasets encompass a wide range of tasks, from basic textual responses to complex multimodal data.
Textual datasets such as MedSTS~\cite{wang2020medsts}, PubMedQA~\cite{jin2019pubmedqadatasetbiomedicalresearch}, and MedQA~\cite{jin2021disease} focus on assessing LLMs on tasks like semantic similarity, question answering, and content summarization in the biomedical domain. 
Specialized datasets like GenBank~\cite{benson2012genbank} test LLMs on their ability to handle genomic sequences, which is crucial for applications in genomics and personalized medicine. 
Multimodal benchmarks like MultiMedBench~\cite{tu2024towards} challenge LLMs to integrate and interpret data from multiple sources, such as medical images and accompanying textual descriptions, reflecting the complex nature of medical diagnostics.
%
%
Evaluation metrics commonly used to assess model performance across different tasks include Accuracy, BLEU-1, F1 Score, and ROUGE-L~\cite{benson2012genbank,hendrycks2021measuringmassivemultitasklanguage,qiu2024towards}. For evaluating LLMs in biomedical dialogue scenarios, specialized metrics such as Professionalism, Fluency, and Safety have been developed to capture the nuanced requirements of biomedical communication~\cite{li2023huatuo26M,wang2023performance,yang2024zhongjing}. 


\begin{table*}[ht]
\centering
\caption{Benchmark datasets and evaluation metrics for evaluating LLMs in the biomedical field.}
\resizebox{1.0\linewidth}{!}{
\begin{tabular}{ccccc}
\hline
\textbf{Dataset} & \textbf{Date} & \textbf{Data Size} & \textbf{Evaluation Metrics}  & \textbf{Description} \\ \hline
MultiMedBench~\cite{tu2024towards} & 2023.07 & $>$1 M & BLEU-1, F1-score  & \makecell[c]{Open-source multimodal biomedical benchmark\\with 14 tasks and 12 de-identified datasets} \\ \hline
GenBank~\cite{benson2012genbank} & 2012.11 & 2 M & F1-score  & \makecell[c]{Public nucleotide sequence database for\\benchmarking} \\ \hline
MedSTS~\cite{wang2020medsts} & 2018.10 & 174,629  & Pearson correlation score  & \makecell[c]{Clinical semantic textual similarity benchmark\\using Mayo Clinic records} \\ \hline
Huatuo26M-test~\cite{li2023huatuo26M} & 2023.05 & 6,000 Q\&A & Professionalism, Fluency, Safety & \makecell[c]{Evaluates single-turn dialogue capability\\in TCM LLMs} \\ \hline
MMLU~\cite{hendrycks2021measuringmassivemultitasklanguage} & 2021.01 & 15,908 Q\&A & Accuracy  & \makecell[c]{Academic benchmark covering 57 subjects\\in English} \\ \hline
PubMedQA~\cite{jin2019pubmedqadatasetbiomedicalresearch} & 2019.09 & 217k Q\&A & Accuracy, ROUGE-L  & \makecell[c]{Biomedical question-answering benchmark\\based on PubMed abstracts} \\ \hline
MedQA~\cite{jin2021disease} & 2021.07 & 10,178 Q\&A & Accuracy, ROUGE-L  & \makecell[c]{Medical question-answering benchmark\\using USMLE exam questions} \\ \hline
MedMCQA~\cite{pal2022medmcqa} & 2022.04 & 194k Q\&A & Accuracy, ROUGE-L  & \makecell[c]{Medical question-answering benchmark\\using Indian entrance exam questions} \\ \hline
MultiMedQA~\cite{singhal2023large} & 2022.12 & 203,282 Q\&A & Accuracy  & \makecell[c]{Comprehensive medical question-answering benchmark\\combining seven datasets} \\ \hline
BioRED~\cite{luo2022biored} & 2022.09 & 20,419  & Precision, Recall, F1-score  & \makecell[c]{Biomedical relation extraction dataset\\for various entity types and relation pairs} \\ \hline
MMedBench~\cite{qiu2024towards} & 2024.02 & 53,566 Q\&A  & ROUGE-1, BLEU-1  & \makecell[c]{Multilingual medical benchmark optimized\\from MMedC} \\ \hline
MacParland~\cite{macparland2018single} & 2018.10 & 8,434  & Accuracy, Macro F1-score  & \makecell[c]{Human liver tissue dataset for new cell type\\detection capability} \\ \hline
CMExam~\cite{liu2024benchmarking} & 2023.06 & 60,000+ Q\&A & Accuracy  & \makecell[c]{Chinese medical comprehensive exam dataset\\for knowledge Q\&A and dialogue} \\ 
ProteinLMBench~\cite{shen2024finetun} & 2024.06 & 944 sixchoice
questions & Accuracy  & \makecell[c]{Protein comprehension} \\
\hline
\end{tabular}}
\label{tab:benchmark_datasets}
\end{table*}

\subsection{Summary}
Our analysis reveals that LLMs, without specialized training, can demonstrate a basic understanding of biomedical terminology and concepts with minimal contextual prompts. 
However, their performance varies across different biomedical disciplines and tasks.
Fig.~\ref{fig:zero-shot}b offers valuable insights into the relative performance of LLMs across different levels of biomedical expertise.
The violin plots indicate that while LLMs generally perform above the baseline across all expertise levels, their performance is most consistent at the intermediate level.
At senior and expert levels, there is greater variability in performance, suggesting that LLMs may struggle with more complex, specialized tasks that require advanced expertise~\cite{meng2024application}.
The evaluation results across various biomedical disciplines highlight both the potential and limitations of LLMs in zero-shot biomedical applications~\cite{liu2023descriptive, ward2022quantitative,horiuchi2024comparing}. 
In certain specific biomedical fields, LLMs show performance comparable to experienced physicians. 
However, in more specialized contexts or complex tasks requiring in-depth biomedical knowledge and clinical reasoning, LLMs may exhibit deficiencies or fail completely.
For most biomedical application scenarios, the zero-shot performance of LLMs falls short of the requirements for immediate clinical application, particularly in highly challenging tasks such as rare disease diagnosis or complex surgical planning~\cite{qiu2023transfer,du2024generative}.
These findings underscore the need for caution when considering the direct application of LLMs to challenging biomedical tasks without fine-tuning or retraining.
While the prospects of LLMs in the biomedical field are promising, it is important to consider their limitations in biomedical applications and thoughtfully define their role in ethical and clinical decision-making processes.
\section{Adapting General LLMs to the Biomedical Field}

\begin{table*}[ht]
\centering
\caption{Overview of large language models in biomedicine.}
\resizebox{1.0\linewidth}{!}{\begin{tabular}{cccccccc}
\hline
\textbf{Model} & \textbf{Date} & \textbf{Parameters} & \textbf{Base Model} & \textbf{Fine-tuning} & \textbf{Tasks and Purpose Description} & \textbf{Unimodal}& \textbf{Open Source} \\
\hline
GatorTron~\cite{yang2022large} & 2022.12 & 8.9B/3.9B/345M & BERT & From scratch & Clinical NLP tasks & \checkmark  & \checkmark\\
BianQue~\cite{chen2023bianque} & 2023.12 & 6B & ChatGLM & Full parameter & Health advice, multi-turn dialogue & \checkmark & \checkmark\\
ChatDoctor~\cite{li2023chatdoctor} & 2023.06 & 7B & LLaMA-7B & Instruction tuning & Medical dialogue & \checkmark & \checkmark\\
DISC-MedLLM~\cite{bao2023disc} & 2023.08 & 13B & Baichuan-13B-Base & Supervised fine-tuning & Medical consultation & \checkmark & \checkmark\\
DNABERT-S~\cite{zhou2024dnabertslearningspeciesawaredna} & 2024.02 & - & BERT & - & DNA sequence analysis & \checkmark & \checkmark\\
GeneCompass~\cite{yang2023genecompass} & 2023.09 & $>$100M & T5 & From scratch & Genomic data analysis & \checkmark & \checkmark\\
GenePT~\cite{chen2023genept} & 2024.03 & - & - & - & Gene and cell representation & \checkmark & \checkmark \\
BenTsao~\cite{wang2023huatuo} & 2023.04 & 7B & LLaMA & Instruction tuning & Chinese biomedical tasks & \checkmark & \checkmark\\
HuatuoGPT~\cite{zhang2023huatuogpt} & 2023.05 & 7B & Bloomz-7b1-mt & Supervised fine-tune, RLAIF & Medical exams, research queries & \checkmark & \checkmark\\
Med-PaLM~\cite{singhal2023large} & 2022.12 & 540B & Flan-PaLM & Prompt tuning & Medical knowledge evaluation & \checkmark & -\\
MedChatZH~\cite{tan2024medchatzh} & 2024.03 & 7B & BaiChuan & Prompt tuning & Chinese medical dialogue & \checkmark & \checkmark\\
Radiology-GPT~\cite{liu2024radiologygptlargelanguagemodel} & 2024.03 & 7B & Alpaca-7B &Instruction tuning, LoRA & Radiology report generation & \checkmark & \checkmark\\
RadOnc-GPT~\cite{liu2023radoncgptlargelanguagemodel} & 2023.11 & - & LLaMA2 & Instruction tuning, LoRA & Radiation treatment planning & \checkmark &- \\
scBERT~\cite{yang2022scbert} & 2022.09 & - & BERT & From scratch & Single-cell RNA analysis & \checkmark & \checkmark\\
scGPT~\cite{cui2024scgpt} & 2024.02 & - & Transformer & From scratch & Single-cell multi-omics analysis & \checkmark & \checkmark\\
Taiyi~\cite{luo2024taiyi} & 2024.02 & 7B & Qwen-7B-base & Supervised fine-tuning & Multilingual biomedical NLP & \checkmark & \checkmark\\
OARA~\cite{guthrie2024operating} & 2024.02 & 7B & Vicuna v1.5 & LoRA & Surgical/anesthetic education & \checkmark &-\\
Med-PaLM 2~\cite{singhal2023towards} & 2023.05 & 340B & PaLm2 & Instruction tuning, LoRA & Advanced medical Q\&A & \checkmark & \checkmark\\
Hypnos~\cite{wang2024towards} & 2024.03 & 7B & LLaMA & LoRA & Anesthesiology tasks & \checkmark &- \\
VetLLM~\cite{jiang2023vetllm} & 2023.12 & 7B & Alpaca-7B & LoRA & Veterinary diagnosis & \checkmark & \checkmark\\
BioMedLM~\cite{bolton2024biomedlm} & 2024.03 & 2.7B & GPT-2 & From scratch & Biomedical Q\&A & \checkmark & \checkmark \\
CancerGPT~\cite{li2024cancergpt} & 2023.04 & 124M & GPT & K-SHOT & Drug synergy prediction & \checkmark &- \\

ESM-2~\cite{lin2023evolutionary} & 2023.03 & 15B & - & From scratch & Protein structure prediction & \checkmark & \checkmark\\
HuatuoGPT II
~\cite{chen2023huatuogptiionestagetrainingmedical} & 2023.11 & 7/13B & Baichuan2-7/13B-Base & Instruction tuning & TCM tasks & \checkmark & \checkmark\\
DoctorGLM~\cite{xiong2023doctorglm} & 2023.04 & 6B & ChatGLM & LoRA & Chinese medical Q\&A & \checkmark & \checkmark\\
MedCPT~\cite{jin2023medcpt} & 2023.11 & - & - & - & Biomedical information retrieval & \checkmark & \checkmark\\
BioGPT~\cite{luo2022biogpt} & 2023.04 & - & GPT-2 & From scratch & Biomedical text generation & \checkmark & \checkmark\\
GatorTronGPT~\cite{peng2023study} & 2023.11 & 5B/20B & GPT-3 & From scratch & Medical text synthesis & \checkmark & \checkmark\\
MEDITRON~\cite{chen2023meditron} & 2023.11 & 7B/70B & LLaMA-2 & Instruction tuning & Medical text comprehension & \checkmark & \checkmark\\
ClinicalGPT~\cite{wang2023clinicalgptlargelanguagemodels} & 2023.06 & 7B & BLOOM-7B & LoRA & Clinical tasks & \checkmark &-\\
Qilin-Med~\cite{ye2024qilinmedmultistageknowledgeinjection} & 2024.04 & - & Baichuan & - & Multi-stage medical training & \checkmark &- \\
MedAlpaca~\cite{han2023medalpacaopensourcecollection} & 2023.01 & 7/13B & LLaMA & LoRA & Open-source medical LLM & \checkmark &-\\
Alpacare~\cite{zhang2023alpacare} & 2024.05 & - & - & Instruction tuning & Medical instruction following & \checkmark & \checkmark\\
Zhongjing~\cite{yang2024zhongjing} & 2023.12 & 13B & Ziya-LLaMA-13B-v13 & Supervised fine-tuning & TCM Q\&A & \checkmark & \checkmark\\
Cpllm~\cite{shoham2023cpllm} & 2024.05 & 13B/2.7B & LLaMA2, PubMedGPT & LoRA & Clinical prediction & \checkmark & \checkmark\\
MMedLM 2~\cite{qiu2024towards} & 2024.02 & 7B & InternLM & LoRA & Multilingual medical Q\&A & \checkmark & \checkmark\\
AlphaFold 3~\cite{abramson2024accurate} & 2024.05 & - & - & - & Protein structure prediction & \checkmark &-\\
\hline
Bingo~\cite{ma2024bingo} & 2023.11 & 15B & ESM-2 & From scratch & Protein-coding gene prediction &  & \checkmark\\
BiomedCLIP~\cite{zhang2024biomed} & 2024.01 & - & CLIP & - & Multimodal biomedical tasks &  & \checkmark\\
Med-PaLm M~\cite{tu2024towards} & 2023.07 & 12B/84B/562B & Palm-E & Instruction tuning & Multimodal medical analysis &  & \checkmark\\
MONET~\cite{kim2024transparent} & 2024.04 & - & CLIP & From scratch & Medical image annotation &  & \checkmark\\
XrayGPT~\cite{thawkar2023xraygptchestradiographssummarization} & 2023.06 & - & MedCLIP, Vicuna & Modality alignment & Chest X-ray analysis &  & \checkmark\\
Med-MLLM~\cite{liu2023medical} & 2023.12 & - & - & Multi-stage training & X-ray representation learning & &- \\
EchoCLIP~\cite{christensen2024vision} & 2024.04 & - & OpenCLIP & From scratch & Echocardiogram interpretation &  & \checkmark\\
OphGLM~\cite{gao2023ophglm} & 2023.06 & 6B & ChatGLM & Instruction tuning & Ophthalmology diagnosis &  & \checkmark\\
ClinicalBLIP~\cite{ji2024vision} & 2024.02 & 3B & InstructBLIP & LoRA & Radiology report generation &  &-\\
Med-Gemini~\cite{saab2024capabilities} & 2024.04 & - & Gemini & Instruction tuning & Multimodal medical analysis & &- \\
BioMedGPT~\cite{luo2023BioMedGPT} & 2023.08 & Instruction tuning & LLaMA2 & - & Biomedical question answering & &\checkmark \\
\hline
\end{tabular}
}
\label{tab:biomedical_llms}
\end{table*}

\begin{figure}[!ht]
\centering
\includegraphics[width=\linewidth]{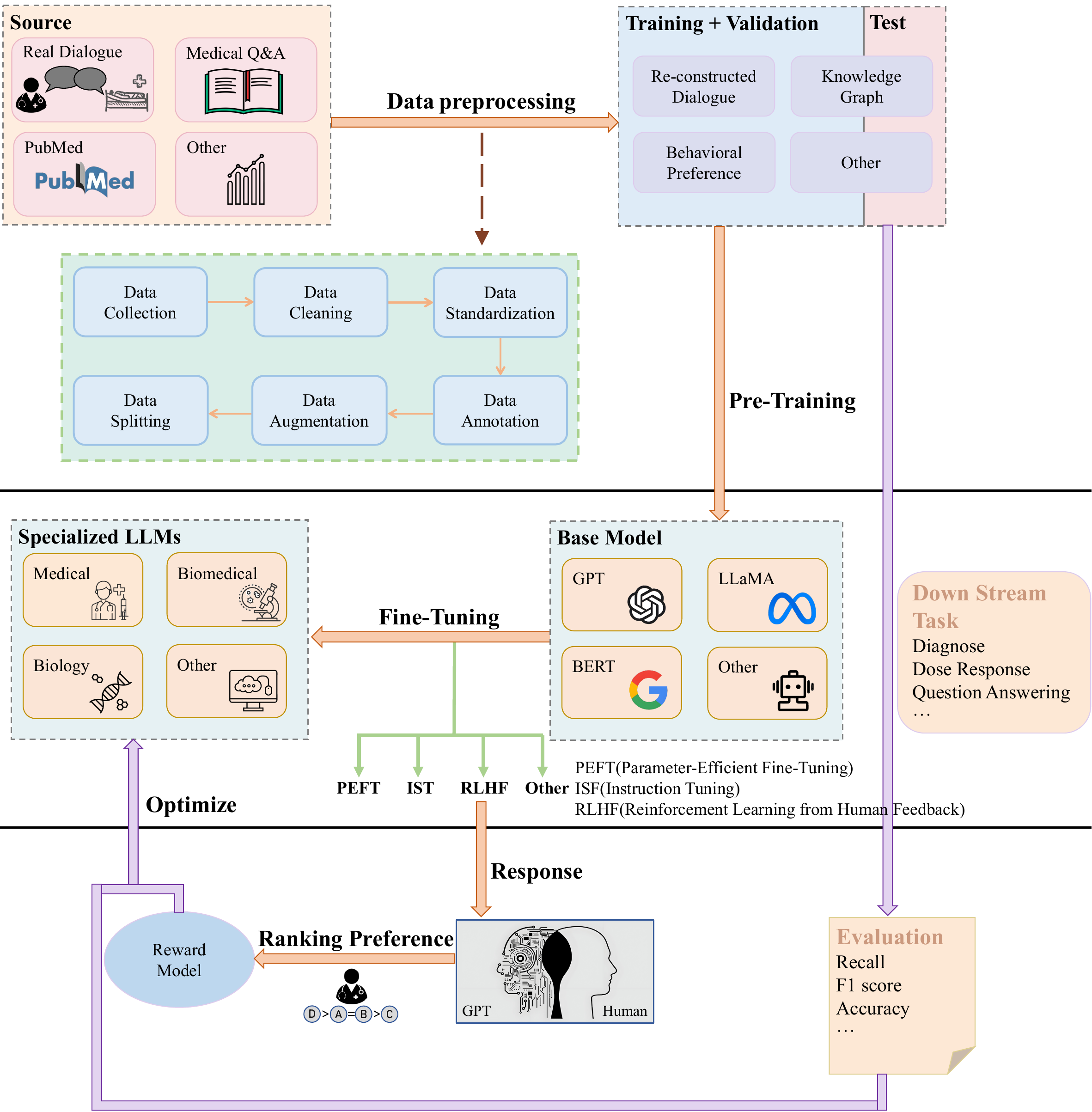}
\caption{Framework for developing and adapting LLMs in biomedicine.
This diagram illustrates the end-to-end process of creating or fine-tuning LLMs for biomedical applications. 
It encompasses data sourcing (\textit{e}.\textit{g}., real dialogues, medical Q\&A, PubMed), preprocessing stages (collection, cleaning, standardization, annotation, and augmentation), and the division into training, validation, and test sets. 
The workflow showcases various pre-training approaches and base models (GPT, LLaMA, BERT) alongside specialized fine-tuning techniques such as PEFT, IFT, and RLHF.
The resulting biomedical LLMs are optimized for downstream tasks like diagnosis, dose-response prediction, and medical question answering. The framework also incorporates evaluation metrics and a feedback loop for continuous improvement, emphasizing the iterative nature of developing effective biomedical LLMs.}
\label{fig:finetune_process}
\end{figure}

General-purpose LLMs encounter various challenges when applied to the biomedical domain in a zero-shot manner, primarily due to the field's highly specialized nature. 
The biomedical sector employs a distinct vocabulary, nomenclature, and conceptual framework that general LLMs may not comprehend~\cite{shen2024protein}.
This specificity extends beyond mere terminology to encompass complex relationships between biological entities, intricate disease mechanisms, and nuanced clinical contexts.
Additionally, the biomedical field presents a diverse array of tasks, ranging from literature analysis and interpretation of clinical notes to supporting diagnostic decisions and drug discovery processes.
This variety demands LLMs capable of performing a wide spectrum of specialized functions, each requiring domain-specific knowledge and reasoning capabilities~\cite{shen2024tagllmrepurposinggeneralpurposellms, luo2024autom3}. 
Moreover, biomedical research increasingly relies on multimodal data integration, incorporating various data types such as text, images (\textit{e}.\textit{g}., radiology scans, histology slides), and molecular sequences (\textit{e}.\textit{g}., DNA, protein structures)~\cite{nam2024harnessing,shen2024toursyn}. 
Effective processing and synthesis of information from these disparate sources pose additional challenges for LLMs. 
To address these challenges and enhance the suitability of general-purpose LLMs for biomedical applications, several adaptation strategies have been developed.
These include domain-specific fine-tuning, architectural modifications, and the creation of specialized biomedical LLMs from the ground up.
Fig.~\ref{fig:finetune_process} illustrates the process of adapting or creating LLMs for biomedical applications, outlining key stages from data preprocessing and curation to model training, fine-tuning, and evaluation.
The adaptation process involves curating high-quality, domain-specific datasets that capture the nuances of biomedical language and knowledge. 
These datasets are then used to fine-tune existing LLMs or train new models, incorporating techniques such as continued pre-training on biomedical corpora, task-specific fine-tuning, and multi-task learning to improve performance across various biomedical tasks~\cite{wang2023huatuo,luo2024taiyi}.
As a result of these efforts, a variety of specialized LLMs have emerged, each tailored to specific aspects of biomedical research and clinical practice. 
Table~\ref{tab:biomedical_llms} provides an overview of these fine-tuned and purpose-built models, showcasing their diversity and specialization within the biomedical domain.

\subsection{Unimodal Adaptation Strategies}
To adapt general-purpose LLMs to the biomedical field, fine-tuning can enable the models to deeply understand the specialized terminology, complex concepts, and linguistic habits of this domain. 
This enhances their ability to provide more accurate and in-depth analysis and generation when dealing with specialized data such as biomedical texts. 
The fine-tuning methods include full-parameter fine-tuning, instruction fine-tuning, parameter-efficient fine-tuning, and hybrid fine-tuning.

\paragraph{Full-Parameter Fine-Tuning} 
Full-parameter fine-tuning involves updating all parameters of a pre-trained LLM using domain-specific data. 
Unlike traditional fine-tuning methods (\textit{e}.\textit{g}., tuning only the top layers), full-parameter fine-tuning allows each layer of the LLMs to learn task-specific knowledge. 
For instance, GatorTron \cite{yang2022large}, a model fine-tuned on clinical data, achieved an F1 score of 93.01\% in medical question answering, surpassing previous benchmarks by 7.77\%. 
%
%
While full-parameter fine-tuning often yields the best performance, it comes with heavy computational costs. 
For instance, fine-tuning GatorTronGPT-20M~\cite{peng2023study} required more than 268,800 GPU hours on A100 GPUs, making it challenging for resource-constrained environments.

\paragraph{Instruction Fine-Tuning} 
Instruction Fine-Tuning (IFT) is a technique that modifies the underlying instructions of a pre-trained model to optimize its adaptation to specific tasks or domains in the biomedical field \cite{zhang2023instruction}. 
This approach has shown promising results in improving model performance on specialized medical tasks.
%
%
For instance, MEDITRON \cite{chen2023meditron}, a model fine-tuned on LLaMA-2 using IFT, demonstrated an average performance improvement of 1.8\% across various medical benchmarks.
Similarly, AlpaCare \cite{zhang2023alpacare} leveraged a curated set of 52,000 medical instructions to achieve a 30.4\% performance boost on the HeadQA benchmark, showcasing the potential of well-designed instruction sets in enhancing model capabilities.
The primary advantage of IFT lies in its ability to adapt models to specific biomedical domains using relatively less data compared to full-parameter fine-tuning. 
However, the effectiveness of IFT heavily depends on the quality and diversity of the instructions used. 
Poorly designed or biased instructions can lead to inconsistent or unreliable model behavior, potentially compromising the model's utility in critical medical applications.
%

\paragraph{Parameter-Efficient Fine-Tuning} 
Parameter-Efficient Fine-Tuning (PEFT) encompasses a set of techniques designed to improve the performance and training efficiency of LLMs by adjusting a small subset of model parameters \cite{houlsby2019parameter}.
Two prominent PEFT approaches are LoRA (Low-Rank Adaptation)~\cite{hu2021lora} and QLoRA (Quantized LoRA)~\cite{dettmers2023qloraefficientfinetuningquantized}, which work by adding small trainable matrices to the model. 
This allows for task-specific adaptations without modifying the entire model architecture. 
The efficiency of PEFT methods is remarkable, often reducing the number of trainable parameters by 99\% or more while maintaining performance comparable to full fine-tuning. 
For example, MMedLM 2 \cite{qiu2024towards} employed LoRA to achieve competitive performance in multilingual medical question-answering tasks while fine-tuning only a fraction of the model's parameters.
This approach reduces computational requirements, making it feasible to deploy tailored medical AI models in resource-constrained environments such as small hospitals or research laboratories.
However, PEFT methods may face limitations when tasks require substantial modifications to the base model's knowledge, as they primarily focus on adapting existing knowledge rather than introducing entirely new information.
This constraint could potentially impact their effectiveness in highly specialized or rapidly evolving areas of biomedicine.
%

\paragraph{Hybrid Fine-Tuning}
Hybrid fine-tuning is an approach that combines multiple parameter-efficient tuning techniques to enhance model performance and training efficiency while minimizing the introduction of additional parameters.
For example, HuatuoGPT~\cite{zhang2023huatuogpt}, using supervised fine-tuning and RLAIF~\cite{bai2022constitutional}, achieves state-of-the-art results in performing medical consultation among open-source LLMs in terms of GPT-4 evaluation, human evaluation, and medical benchmark datasets. 
Hybrid fine-tuning strategies offer a balance between performance and efficiency, addressing some of the limitations of individual techniques. 
They allow for more flexible adaptation to the unique challenges of medical AI, such as the need for both broad medical knowledge and specialized expertise. 
However, these approaches often require more complex implementation and careful tuning of multiple components. 
%

\subsection{Multimodal Adaptation Strategies}
Multimodal LLMs represent can integrate diverse data types to provide comprehensive insights. 
The core strength of these models lies in their ability to fuse information from various modalities, including text, images, gene sequences, and protein structures. 
This fusion not only bridges interdisciplinary gaps but also mirrors the multifaceted nature of medical diagnosis and research~\cite{zhao2024deep}.
In clinical settings, patient assessments typically involve an array of data types, including textual information (\textit{e}.\textit{g}., medical reports), visual data (\textit{e}.\textit{g}., X-rays and MRIs), and numerical measurements (\textit{e}.\textit{g}., laboratory results and vital signs). 
Multimodal LLMs aim to integrate these diverse sources to offer more accurate and holistic biomedical insights. 
For instance, by combining medical imaging with clinical text reports and other relevant data, these models can improve diagnostic accuracy and robustness~\cite{zhou2023transformer}.
In addition, multimodal can facilitate the integration of genomic data with phenotypic information, enabling more comprehensive studies of disease mechanisms and discover new drugs~\cite{luo2023BioMedGPT}. 

Fine-tuning strategies play a crucial role in the application of biomedical multimodal models, ensuring that these models can adequately comprehend and process cross-modal data. These strategies encompass various approaches, including the optimization of visual encoders through LoRA~\cite{hu2021lora} and layer normalization~\cite{ba2016layernormalization} techniques. Such optimizations are implemented to enhance the model's capacity to interpret critical features within medical images. Concurrently, these strategies integrate visual and textual inputs, leveraging attention mechanisms and multilayer perceptron (MLP) layers to augment the model's proficiency in generating radiology reports, as exemplified by the ClinicalBLIP~\cite{ji2024vision} model.
Specifically, ClinicalBLIP demonstrated superior performance in the radiology report generation task using the MIMIC-CXR~\cite{johnson2019mimic} dataset, achieving a Metric For Evaluation of Translation
with Explicit Ordering (METEOR)~\cite{banerjee2005meteor} score of 0.534 through these fine-tuning strategies. This score significantly surpasses that of other models, underscoring ClinicalBLIP's exceptional capability in handling complex multimodal data.
Similarly, Med-Gemini~\cite{saab2024capabilities} employs a strategy of constructing a joint embedding space, enabling direct comparison and integration of data from diverse modalities within a unified latent space. This approach has exhibited remarkable performance in complex medical tasks, particularly in cancer diagnostics, where the integration of genomic data and pathological images has substantially enhanced diagnostic accuracy.
These fine-tuning strategies, by optimizing model performance in biomedical multimodal tasks, demonstrate the immense potential of applying multimodal models in the medical domain. Furthermore, they underscore the critical role of fine-tuning in enhancing model generalization capabilities and task adaptability.

\subsection{Training Data and Processing Strategies}

The adaptation of general-purpose LLMs to the biomedical domain hinges on the quality, diversity, and processing of the data.
This subsection explores key datasets and effective strategies for developing and refining biomedical LLMs.

\subsubsection{Dataset Overview}
Biomedical datasets utilized for LLM training and evaluation span three main categories, namely text-based, image-based, and multimodal.
Table~\ref{tab:datasets} summarizes datasets employed in recent studies.
Text-based datasets, such as PubMed, have been instrumental in training models like BioGPT~\cite{luo2022biogpt}. 
Similarly, the MIMIC-III dataset, containing de-identified health records from over 40,000 care patients, contributes to models like GatorTron~\cite{yang2022large}, enabling LLMs to learn from real-world clinical data.
Multimodal datasets, which integrate various data types, facilitate more comprehensive model training. 
The MultiMedBench~\cite{tu2024towards} dataset exemplifies this approach by aligning clinical notes with medical measurements and imaging data.
Models like Med-PaLM M~\cite{tu2024towards} trained on such datasets demonstrate enhanced performance in tasks requiring the integration of heterogeneous data types, bridging the gap between textual and visual medical information.

\subsubsection{Data Processing Strategies}

To maximize the utility of these datasets, researchers have employed various data processing techniques.

\paragraph{Data Augmentation}
Augmentations aim to increase dataset size and diversity, thereby improving model robustness and generalization. 
Chen \textit{et al.}~\cite{chen2023bianque}, in their development of BianQue by combining automatic data cleaning with ChatGPT-based data polishing. 
This method not only enhanced the quality of training data but also led to a 15\% improvement in the model's performance on medical consultation tasks.

\paragraph{Data Mixing}
The integration of diverse data sources can also enhance model capabilities. 
Bao \textit{et al.}~\cite{bao2023disc} demonstrated this in DISC-MedLLM, employing a data fusion strategy. 
By combining structured information from medical knowledge graphs with human-curated samples, they achieved a 20\% improvement in handling medical queries compared to models trained on single-source data.
%

\subsubsection{Federated Learning in LLMs}
In the realm of biomedical LLMs, direct data sharing is often impractical due to stringent healthcare regulations. 
Federated Learning (FL) ~\cite{zhang2021survey} has emerged as a transformative solution, potentially reshaping the future of LLM training in healthcare. 
Unlike traditional LLMs trained on single, proprietary data centers, biomedical LLMs require diverse datasets that can be effectively accessed through FL. 
%
%
The OpenFedLLM framework \cite{ye2024openfedllm}, facilitates FL across geographically distributed datasets while promoting ethical alignment. 
Complementing this, Wu \textit{et al.} \cite{wu2020fedmed} introduced FedMed, a framework specifically designed to enhance medical language modeling while mitigating performance degradation in federated settings. 
Zhang \textit{et al.} \cite{zhang2023gpt} further advanced the field by demonstrating the effectiveness of combining FL with prompt-based approaches for clinical applications, enhancing model adaptability while preserving patient privacy. 
Nagy \textit{et al.} \cite{nagy2023privacy} explored privacy-preserving techniques for training large language models like BERT and GPT-3, providing insights into maintaining privacy without compromising performance. 
Addressing multilingual challenges, Weller \textit{et al.} \cite{weller2022pretrained} investigated the use of pre-trained language models in FL across multiple languages, focusing on various NLP tasks in medical contexts. 
Finally, Kim \textit{et al.} \cite{kim2024efficient} proposed improving computational efficiency in FL by integrating adapter mechanisms into pre-trained LLMs, demonstrating the benefits of using smaller Transformer-based models to reduce computational demands. 

\begin{table*}[ht]
\centering
\caption{Datasets for fine-tuning and evaluating biomedical LLMs.}

\resizebox{1.0\linewidth}{!}{
\begin{tabular}{cccc}
\hline
\textbf{Dataset} & \textbf{Date} & \textbf{Data Size} & \textbf{Description}  \\ \hline
MIMIC-CXR~\cite{johnson2019mimic} & 2019.12 & 377,110 chest X-rays and reports & De-identified medical data for training image-text pairs to improve diagnostic accuracy  \\ 
IU X-ray~\cite{demner2016preparing} & 2016.03 & 7,470 images and 3,955 reports & Chest X-rays for training models in interpreting X-ray images and reports  \\ 
COVID-19-CT~\cite{demner2016preparing} & 2021.07 & 1,104 images and 368 reports & COVID-19 CT images and reports for enhancing model analysis of COVID-19 data  \\ 
DDI~\cite{herrero2013ddi} & 2013.07 & 18,502 pharmacological substances and 5028 DDIs  & Clinical images for drug-drug interaction extraction  \\ 
OpenI~\cite{demner2016preparing} & 2015.07 & 6,459 images and 3,955 reports & Chest X-rays for training models in medical image and report interpretation \\ 
VQA-RAD~\cite{lau2018dataset} & 2018.11 & 315 radiology images and 3,515 Q\&A  & Radiology Visual Question Answering dataset  \\ 
Slake-VQA~\cite{liu2021slake} & 2021.02 & 642 images and 14,028 Q\&A  & Bilingual VQA dataset for medical visual question answering  \\ 
Path-VQA~\cite{he2020pathvqa30000questionsmedical} & 2020.03 & 4,998 images and 32,799 Q\&A  & Pathology VQA dataset for understanding pathology images  \\ 
PMC-15M~\cite{zhang2024biomed} & 2024.01 & 15 M & Scientific article data for biomedical image and text analysis  \\ 
ChiMed-CPT~\cite{ye2024qilinmedmultistageknowledgeinjection} & 2024.04 & 2 B & QilinMed: Enhancing medical knowledge in LLMs \\ 
ProteinLMDataset~\cite{shen2024finetun} & 2024.06 & 17.46 B tokens and 893K instructions & Protein sequence comprehension\\
\hline 
scCompass-126M~\cite{yang2023genecompass} & 2023.09 & 126 M & Genomics research data from humans and mice  \\ 
PanglaoDB~\cite{franzen2019panglaodb} & 2019.01 & 209  & Single-cell biology data for the scBERT project  \\ 
CMtMedQA~\cite{yang2024zhongjing} & 2023.12 & 70,000 Q\&A  & Real doctor-patient dialogues for complex medical Q\&A  \\ 
huatuo-26M~\cite{li2023huatuo26M} & 2023.05 & 26 M Q\&A  & Chinese medical dialogues for Q\&A systems  \\ 
BC5CDR~\cite{li2016biocreative} & 2016.04 & 13,343 & PubMed articles for chemical-disease relation extraction  \\ 
HealthSearchQA~\cite{singhal2023large} & 2022.12 & 3,375 Q\&A  & Data for answering common health search queries  \\ 
cMedQA2~\cite{singhal2023large} & 2018.12 & 120,000 Q\&A & Consumer medical questions dataset  \\ 
MedDialog~\cite{zhang2018multi} & 2020.11 & 5.1 M & Chinese medical Q\&A dataset  \\ 
BianQueCorpus~\cite{chen2023bianque} & 2023.12 & 2,437,190 & Multi-turn medical dialogues from online platforms  \\ 
MIMIC-III~\cite{johnson2016mimic} & 2016.05 & 5 B & Optimized dialogues for health-related ChatGPT training  \\ 
webmedQA~\cite{he2019applying} & 2018.12 & 63,284 Q\&A & Clinical domain corpus for question answering  \\ 
MedInstruct-52k~\cite{zhang2023alpacare} & 2024.05 & 52,000& Dataset for medical instruction-following tasks  \\ \hline

\end{tabular}
}
\label{tab:datasets}
\end{table*}

\subsection{Summary}
This section has explored the adaptation of general-purpose LLMs to the biomedical domain, highlighting the important interplay between data quality, processing strategies, and model adaptation techniques. 
We reviewed the foundational role of diverse datasets and advanced data processing methods in developing robust biomedical LLMs. 
The investigation of various adaptation approaches, from full-parameter fine-tuning to more efficient methods like instruction tuning and parameter-efficient techniques.
Despite these advancements, challenges persist in data privacy, model interpretability, and fairness.
Future research can focus on developing more efficient, interpretable, and ethical adaptation techniques. 
Priority areas include enhancing model transparency, addressing fairness concerns, and exploring advanced federated learning methods to leverage decentralized medical data while preserving patient privacy. 
The integration of multimodal approaches also presents a promising avenue for more comprehensive healthcare solutions.
As biomedical LLMs continue to evolve, balancing technological innovation with ethical considerations will be important. 
By addressing current challenges and embracing emerging opportunities, these models have the potential to revolutionize healthcare, from improving clinical decision support to accelerating biomedical research, ultimately leading to more effective and equitable healthcare delivery.

\section{Discussion}
\subsection{Challenges of LLMs in Biomedical Applications}
LLMs have demonstrated potential in biomedical applications, as evidenced by our review of zero-shot evaluations and adaptation strategies. 
While unadapted LLMs show promise in certain tasks, fine-tuning has proven crucial in bridging the gap between general language understanding and specialized medical knowledge.
Unimodal LLMs, after appropriate adaptation, have achieved improvements in processing medical texts, answering complex questions, and facilitating medical dialogues.
For example, GatorTron excelled in various clinical NLP tasks after full-parameter fine-tuning~\cite{yang2022large}, while MMedLM 2 demonstrated competitive performance in multilingual medical question answering using parameter-efficient fine-tuning methods~\cite{qiu2024towards}.
Multimodal LLMs have expanded the horizons of medical diagnosis and analysis by integrating image and text data. 
Models such as Med-Gemini~\cite{saab2024capabilities} and Med-PaLM M~\cite{tu2024towards} have shown promising results in tasks requiring the integration of visual and textual information, enhancing the accuracy of medical imaging processing and diagnosis.

Compared to traditional machine learning methods in biomedicine, LLMs offer several advantages, including improved generalization across tasks and enhanced performance on complex reasoning tasks. 
However, they also face challenges including higher computational requirements and the need for large, diverse datasets for effective training and adaptation.
Data privacy and security concerns remain paramount when handling sensitive patient information. 
The lack of interpretability in LLM decision-making processes raises trust and accountability issues in clinical settings. 
The quality and diversity of training datasets significantly impact model performance and generalizability, while the substantial computational resources required for training and fine-tuning limit widespread application, particularly in resource-constrained environments. 
Additionally, ethical considerations surrounding potential biases in training data and model outputs necessitate careful scrutiny and mitigation strategies.

\subsection{LLMs Across Healthcare Hierarchy}
LLMs demonstrate potential in healthcare, yet their practical implementation necessitates careful consideration of the hierarchical structure within medical systems. 
The role and impact of LLMs vary across different levels of healthcare delivery, from high-level management to primary care~\cite{maleki2024clinicaltrialsprotocolauthoring}.
At the administrative level, LLMs have the potential to improve the decision-making processes by analyzing vast data to optimize resource allocation and forecast healthcare demands.
For specialist physicians, these models can serve as powerful diagnostic adjuncts, integrating the latest research findings to inform personalized treatment recommendations. 
In routine clinical practice, LLM-augmented intelligent triage systems and medical image interpretation tools hold promise for enhancing the diagnostic efficiency and accuracy of junior doctors.
Of major importance is the potential of LLMs to ameliorate primary healthcare, especially in resource-constrained settings. 
In underserved areas, lightweight LLM models could provide basic diagnostic support, while telemedicine platforms powered by these models could bridge the urban-rural healthcare divide by connecting disparate medical resources.
However, the integration of LLMs into medical practice faces multifaceted challenges. 
Model customization to specific medical specialties and local healthcare contexts is important, as is ensuring continuous updating to keep models current with the latest medical knowledge and practices. 
Ethical considerations, including addressing issues of bias, privacy, and transparency in LLM-assisted decision-making, must be at the forefront of implementation efforts. 
Rigorous clinical validation against established medical standards and comprehensive user training for healthcare professionals on the appropriate use and limitations of LLM tools are also essential steps in the integration process.


\begin{figure}[!ht]
    \centering
    \includegraphics[width=\linewidth]{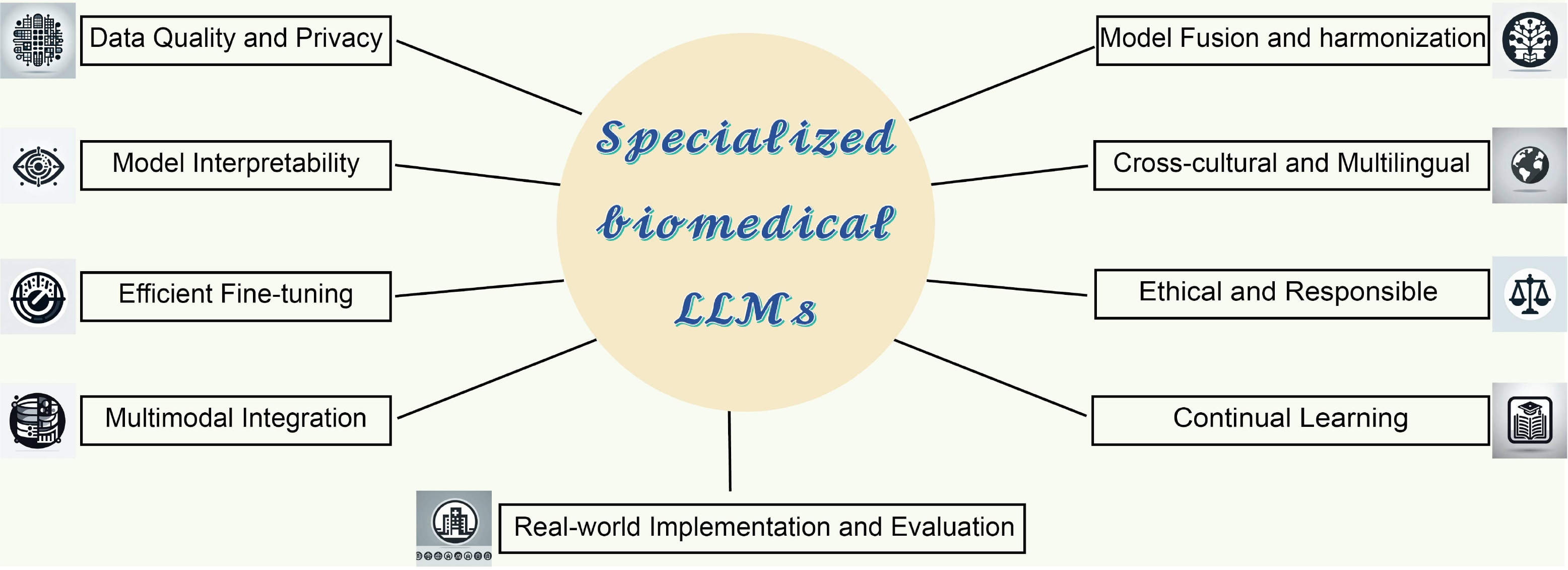}
    \caption{Future directions of LLMs in the biomedical field.
    }
    \label{fig:future}
\end{figure}
\subsection{Future Direction}
The integration of LLM in biomedicine presents opportunities alongside important ethical considerations. 
These include potential algorithmic bias, informed consent in AI-assisted clinical decision-making, medical responsibility, and liability issues, and concerns about data ownership and privacy. 
Addressing these challenges requires ongoing collaboration between AI researchers, healthcare professionals, ethicists, and policymakers to develop robust guidelines and regulatory frameworks.

Future research directions in this field are multifaceted and interconnected (Fig.~\ref{fig:future}). 
Enhancing data quality and diversity through interdisciplinary collaboration is important for improving model performance and reducing biases. 
In this context, emerging techniques such as FL and differential privacy offer promising solutions to data privacy concerns while maintaining model performance~\cite{wang2023ppefl}. 
Simultaneously, developing more interpretable models and user-friendly interfaces can increase trust and adoption in clinical settings.
Techniques such as attention visualization, concept attribution, and local interpretable model-agnostic explanations (LIME)~\cite{ribeiro2016should} can be further explored and adapted for biomedical LLMs.
The exploration of efficient fine-tuning methods, particularly parameter-efficient techniques, holds promise for enhancing the applicability and performance of LLMs across various medical specialties while reducing computational costs. 
%
Model fusion and harmonization represent an important frontier in biomedical AI~\cite{li2023deepmodelfusionsurvey}. 
Future research should focus on developing advanced techniques for combining multiple specialized LLMs to create more comprehensive and robust systems.
This approach holds promise for addressing the complex, multifaceted nature of medical knowledge and decision-making.
The cross-cultural adaptability of LLMs is essential for ensuring their global applicability in diverse healthcare systems. 
This challenge calls for the development of multilingual models capable of understanding and generating medical content across languages and cultural contexts, which is important for bridging healthcare disparities and ensuring equitable access to AI-powered medical support worldwide.
Continued research into ethical AI practices specific to biomedical applications is also important. This encompasses developing frameworks for fair and unbiased model development, ensuring informed consent in AI-assisted clinical decision-making, and establishing clear guidelines for the responsible use of LLMs in healthcare. 
Additionally, future research can also focus on implementing LLMs in real-world clinical settings and conducting rigorous evaluations of their performance, impact on patient outcomes, and integration with existing healthcare workflows.
Lastly, the rapid evolution of medical knowledge necessitates the development of methods for continual learning and adaptation of LLMs. 
This ongoing refinement is crucial to ensure that these models remain at the forefront of medical knowledge and practice, capable of incorporating new discoveries and changing treatment paradigms in real time.

\section{Conclusion}
In this study, we have explored the potential and applications of general-purpose LLMs in the biomedical field. By evaluating the performance of unimodal and multimodal LLMs in processing medical texts, images, and integrated data, we have validated the potential of these LLMs in enhancing the efficiency and accuracy of medical research. Our research first provided an overview of the current state of LLMs in the biomedical field, highlighting the limitations of directly applying general LLMs and emphasizing the importance of fine-tuning strategies. Despite the broad application prospects of LLMs, their application in the biomedical field faces several challenges, including data privacy and security, model interpretability, dataset quality and diversity, and high computational resource demands. These challenges limit the widespread application of LLMs. To address these challenges, we proposed future directions including improving data quality and diversity, enhancing model interpretability, developing efficient and economical fine-tuning methods, exploring multimodal data fusion techniques, and promoting interdisciplinary collaboration. These measures will further advance the application and development of LLMs in the biomedical field.


\bibliography{main}
\end{document}